\documentclass[runningheads]{llncs}

\usepackage{eccv}

\usepackage{eccvabbrv}

\usepackage{graphicx}
\usepackage{booktabs}

\usepackage[accsupp]{axessibility}  

\usepackage{hyperref}

\usepackage{orcidlink}

\newcommand{\figureNegMining}{
\begin{figure}[tp] 
	\centering
	\includegraphics[width=0.95\textwidth]{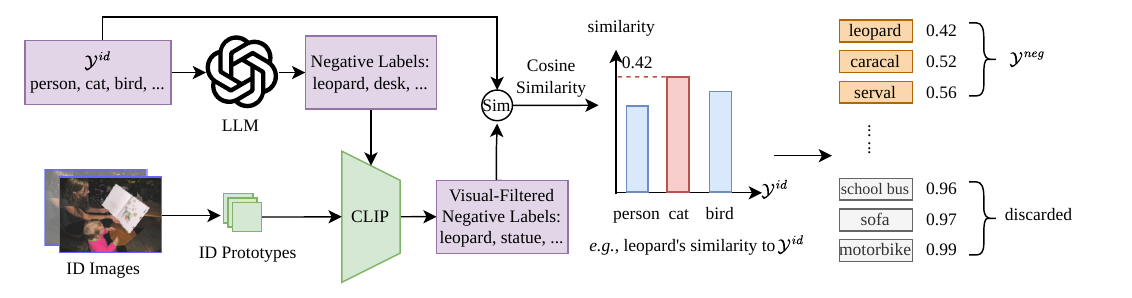}
	\caption{Negative Label Mining. Given ID category texts, we prompt an LLM to generate candidate negative labels. Each candidate is compared to every ID label by computing their cosine similarity in the text embedding space. The candidates are sorted by their minimum similarity to the ID label set, and the $K$ labels with the lowest minimum similarity are selected to form $\mathcal{Y}^{\text{neg}}$. Labels with high similarity (\eg, ``school bus'', ``sofa'') are discarded to avoid semantic overlap with ID categories. 
    }
	\label{fig:negmining} 
\end{figure}
}

\newcommand{\figureMethodPipeline}{
\begin{figure*}[tp] 
	\centering
	\includegraphics[width=1.0\textwidth]{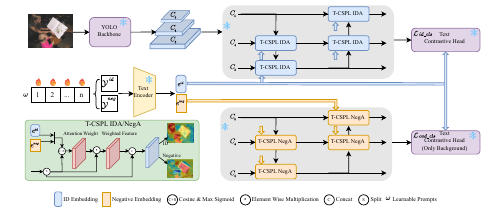}
	\caption{The framework of our method. Based on the original YOLO-World, we add an additional NegA-based OOD branch. NegA leverages the interaction between the text embeddings of LLM-curated negative labels and visual features to enhance potential OOD background regions. The enhanced features are then fed into the detection head to compute the classification loss for the background regions only. Through backpropagation, the learnable prompts in the text encoder are updated, helping the model learn a better decision boundary. During inference, the NegA attention branch is discarded (no additional inference cost), while the negative text embeddings are retained for computing the NegS OOD score.}
	\label{fig:oodattn} 
\end{figure*}
}

\newcommand{\figureHeatmap}{
	\begin{figure}[htp]
		\centering
		\includegraphics[width=0.8\linewidth]{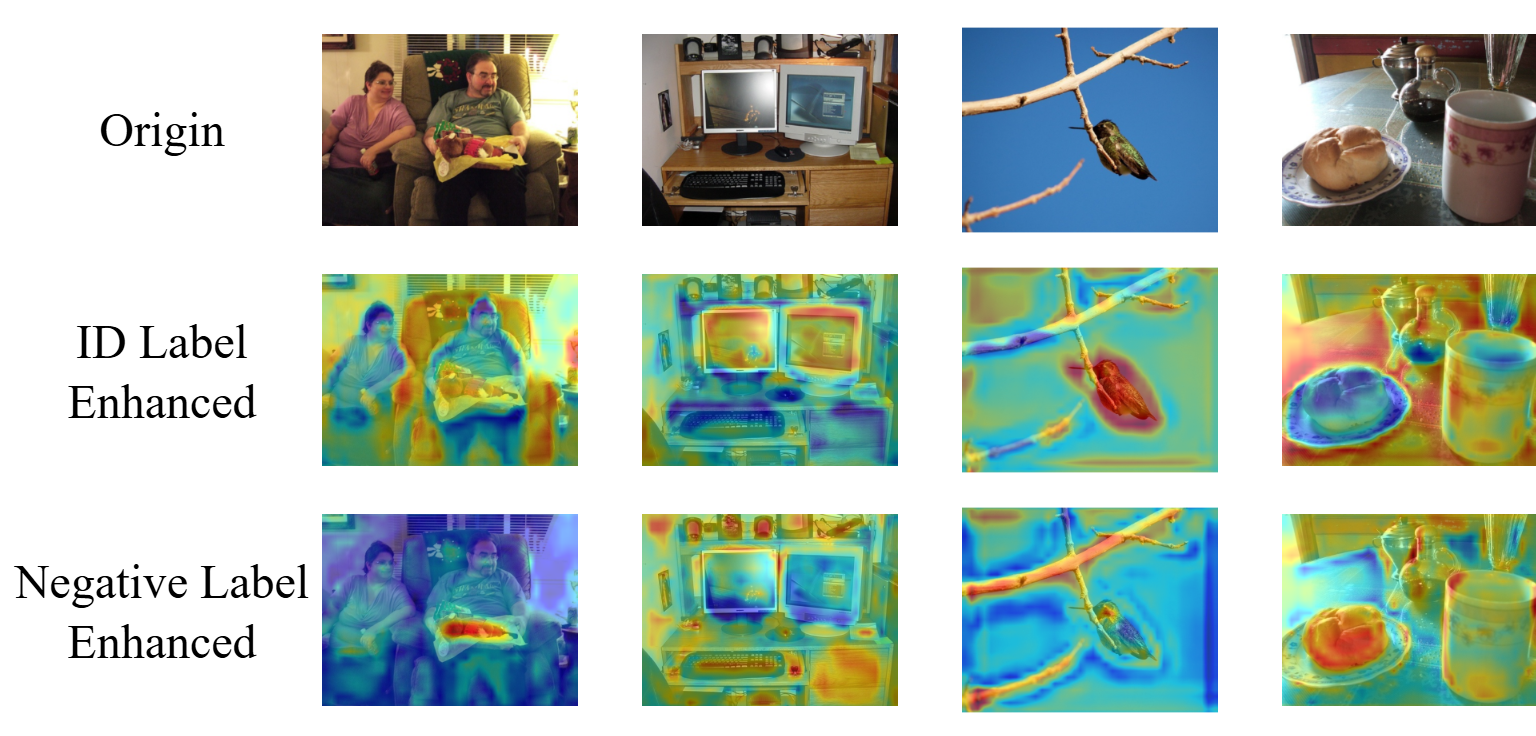}
		\caption{
			Feature heatmap comparison between YOLO-World and NegAS.
			The feature maps are extracted from the neck layer.
		}
		\label{fig:heatmap_visualization}
	\end{figure}
}

\newcommand{\tableComparisonBaselines}{
	\begin{table}[!ht]
		\centering
		\caption{Comparison with baseline models on OOD detection performance on different in-distribution datasets. 
			Best results are in \textbf{bold}.
			``$\text{YOLO-World}$'' and ``$\text{GroundingDINO}$'' means the original YOLO-World and GroundingDINO models without modification.  
			``$\text{YOLO-World}_{CoOp}$'' denotes incorporating CoOp learnable prompts into YOLO-World and training only prompts on ID labels.
			``$\text{NegAS}$'' denotes our proposed method. 
            $\uparrow$ ($\downarrow$) denotes performance improvements compared with the baseline.
            ``YW'' means YOLO-World and ``GD'' means Grounding DINO.
            Note that the mAP (ID) gain over YW is largely attributable to CoOp prompt tuning; NegAS \emph{preserves} this ID accuracy (\cf~\Cref{tab:table_id_map_std}) while substantially improving OOD detection.
		}
		\label{tab:table_comparison_baselines}
        \scalebox{0.9}{
        \setlength{\tabcolsep}{1mm}
        \begin{tabular}{c|c|l|ccccc}
			\toprule
			& & & \multicolumn{4}{c}{OOD Dataset} & \\
		       & & \multirow{2}{*}{Method} & 
			\multicolumn{2}{c}{MS-COCO} & 
			\multicolumn{2}{c}{OpenImages} & 
			\multirow{2}{*}{mAP (ID)$\uparrow$} \\
			\cmidrule(lr){4-5} \cmidrule(lr){6-7}
			& & & FPR95$\downarrow$ & AUROC$\uparrow$ & FPR95$\downarrow$ & AUROC$\uparrow$ &  \\
			\midrule
			\multirow{17}{*}{\rotatebox{90} {VOC}} & \multirow{12}{*}{\rotatebox{90}{Traditional Detector}}
            & Energy~\cite{liu2020energy} & 56.9 & 83.7 & 58.7 & 83.0 & - \\
            & & Gram Matrices~\cite{sastry2020detecting} & 62.8 & 79.9 & 67.4 & 48.7 & 60.6 \\
            & & Generalized ODIN~\cite{hsu2020generalized} & 59.6 & 83.1 & 70.3 & 79.2 & - \\
            & & CSI~\cite{tack2020csi} & 59.9 & 81.8 & 57.4 & 83.0 & 59.5 \\
            & & Siren-vMF~\cite{du2022siren} & 75.5 & 76.1 & 78.4 & 71.1 & 60.8 \\
            & & Siren-KKN~\cite{du2022siren} & 64.8 & 78.2 & 66.0 & 74.9 & 60.8 \\
            & & SR-VAE~\cite{wu2023discriminating} & 42.2 & 90.3 & 46.3 & 87.9 & - \\
            & & DFFF~\cite{wu2023deep} & 41.3 & 90.8 & 44.5 & 88.7 & - \\
            & & SyncOOD~\cite{liu2024can} & 36.4 & 86.5 & 13.3 & \textbf{95.4} & - \\
            & & WFS~\cite{wu2025percept} & 37.5 & 85.6 & \textbf{11.3} & 94.9 & - \\
            & & VOS~\cite{du2022vos} & 47.5 & 88.7 & 51.3 & 85.2 & 60.3 \\
            & & SAFE~\cite{wilson2023safe} & 47.4 & 80.3 & 20.1 & 92.3 & 60.8 \\
            \cmidrule{2-8}
            
            & \multirow{5}{*}{\rotatebox{90}{VLM}} & $\text{YW}$(baseline)~\cite{cheng2024yolo} &  21.3 & 93.4 & 48.8 & 81.4 & 69.0 \\
			& & $\text{YW}_{CoOp}$(baseline) & 13.9 & 95.7 & 47.0 & 83.3 & 72.3 \\
			& & \cellcolor{gray!20}$\text{YW + NegAS}$ &
			\cellcolor{gray!20}\textbf{9.9}$_{\textcolor{mycolor5}{\downarrow 11.4}}$ & \cellcolor{gray!20}\textbf{95.9}$_{\textcolor{mycolor5}{\uparrow 1.5}}$ & 
			\cellcolor{gray!20} {23.3}$_{\textcolor{mycolor5}{\downarrow 25.5}}$ & \cellcolor{gray!20} 92.6$_{\textcolor{mycolor5}{\uparrow 11.2}}$ & 
			\cellcolor{gray!20} \textbf{72.5}$_{\textcolor{mycolor5}{\uparrow 3.5}}$ \\
            
			& & $\text{GD}$(baseline)~\cite{liu2024grounding} &  11.7 & 96.8 & 33.4 & 91.8 & 63.8 \\
            
			& & \cellcolor{gray!20}$\text{GD+NegAS}$ & 
			\cellcolor{gray!20}\textbf{7.2}$_{\textcolor{mycolor5}{\downarrow 4.5}}$ & \cellcolor{gray!20}\textbf{97.6}$_{\textcolor{mycolor5}{\uparrow 0.8}}$ & 
			\cellcolor{gray!20} {24.3}$_{\textcolor{mycolor5}{\downarrow 9.1}}$ & \cellcolor{gray!20} 92.1$_{\textcolor{mycolor5}{\uparrow 0.3}}$ & 
			\cellcolor{gray!20} \textbf{66.1}$_{\textcolor{mycolor5}{\uparrow 2.3}}$ \\
            
			\midrule
			\multirow{17}{*}{\rotatebox{90} {BDD}} & \multirow{12}{*}{\rotatebox{90}{Traditional Detector}}
            & Energy~\cite{liu2020energy} & 60.1 & 77.5 & 55.0 & 80.0 & - \\
            & & Gram Matrices~\cite{sastry2020detecting} & 60.9 & 74.9 & 77.6 & 59.4 & 30.3 \\
            & & Generalized ODIN~\cite{hsu2020generalized} & 57.3 & 85.2 & 50.2 & 87.2 & - \\
            & & CSI~\cite{tack2020csi} & 47.1 & 84.1 & 37.1 & 88.0 & 29.9 \\
            & & Siren-vMF & 67.5 & 80.1 & 66.3 & 79.8 & 31.3 \\
            & & Siren-KKN & 54.0 & 86.6 & 47.3 & 89.0 & 31.3 \\
            & & SR-VAE~\cite{wu2023discriminating} & 32.2 & 90.7 & 21.8 & 93.6 & - \\
            & & DFFF~\cite{wu2023deep} & 30.7 & 90.7 & 22.7 & 92.5 & - \\
            & & SyncOOD~\cite{liu2024can} & 22.7 & 95.4 & 13.0 & 96.3 & - \\
            & & WFS~\cite{wu2025percept} &  21.8 & 93.4 & 7.8 & 96.9& - \\
            & & VOS~\cite{du2022vos} &  44.3 & 86.9 & 35.5 & 88.5 & 31.1 \\
            & & SAFE~\cite{wilson2023safe} &  32.6 & 89.0 & 16.0 & 94.6 & \textbf{31.2} \\
            \cmidrule{2-8}
            
            & \multirow{5}{*}{\rotatebox{90}{VLM}} & $\text{YW}$(baseline)~\cite{cheng2024yolo} &  40.4 & 83.9 & 8.2 & 97.0 & 22.5 \\
			& & $\text{YW}_{CoOp}$(baseline) &  47.0 & 82.0 & 15.1 & 92.6 & 26.1 \\
			& & \cellcolor{gray!20}$\text{YW + NegAS}$ &
			\cellcolor{gray!20}\textbf{20.9}$_{\textcolor{mycolor5}{\downarrow 19.5}}$ & \cellcolor{gray!20}\textbf{92.0}$_{\textcolor{mycolor5}{\uparrow 9.0}}$ & 
			\cellcolor{gray!20}\textbf{3.6}$_{\textcolor{mycolor5}{\downarrow 5.4}}$ & \cellcolor{gray!20}\textbf{98.7}$_{\textcolor{mycolor5}{\uparrow 1.7}}$ & 
			\cellcolor{gray!20}29.0$_{\textcolor{mycolor5}{\uparrow 6.5}}$ \\
            
			& & $\text{GD}$(baseline)~\cite{liu2024grounding} & 21.2 & 94.4 & 21.6 & 94.2 & 23.6 \\
            
			& & \cellcolor{gray!20}$\text{GD+NegAS}$ & 
			\cellcolor{gray!20}\textbf{17.8}$_{\textcolor{mycolor5}{\downarrow 3.4}}$ & \cellcolor{gray!20}\textbf{95.8}$_{\textcolor{mycolor5}{\uparrow 1.4}}$ & 
			\cellcolor{gray!20} \textbf{11.5}$_{\textcolor{mycolor5}{\downarrow 10.1}}$ & \cellcolor{gray!20}\textbf{95.2}$_{\textcolor{mycolor5}{\uparrow 1.0}}$ & 
			\cellcolor{gray!20} 25.2$_{\textcolor{mycolor5}{\uparrow 1.6}}$ \\
            
			\bottomrule
		\end{tabular}
        }
	\end{table}
}

\newcommand{\tableAblationComparison}{
	\begin{table}[tbp]
		\centering
		\caption{
		Ablation study on the effectiveness of the proposed NegA and NegS components. 
		All variants are trained with learnable prompts.
		\ding{51} and \ding{55} denote the presence and absence of each component, respectively.
		}
		\label{tab:table_ablation_comparison}
        \scalebox{0.8}{
        \setlength{\tabcolsep}{1.5mm}
		\begin{tabular}{l|cc|cc}
			\toprule
			\multirow{2}{*}{ID $\mathcal{D}$} & \multirow{2}{*}{NegA} & \multirow{2}{*}{NegS}& \multicolumn{2}{c}{OOD: MS-COCO / OpenImages}   \\
			\cmidrule(lr){4-5}
			& & &FPR95$\downarrow$ & AUROC$\uparrow$  \\
			\midrule
			\multirow{4}{*}{VOC} 
			& \ding{55}  & \ding{55}     & 13.9 / 47.0 & 95.7 / 83.3\\
			& \ding{55}  & \ding{51}     & 11.4 / 33.3 & 96.2 / 88.8 \\
			& \ding{51} & \ding{55}  & 11.8 / 41.8 & \textbf{96.4} / 84.4  \\
			& \ding{51} & \ding{51} & \textbf{9.9} / \textbf{23.3} & 95.9 / \textbf{92.6}  \\
			\midrule
			\multirow{4}{*}{BDD} 
			& \ding{55}  & \ding{55} & 47.0 / 15.1 & 82.0 / 92.6 \\
			& \ding{55}  & \ding{51} & 43.6 / 12.2 & 82.9 / 94.5 \\
			& \ding{51} & \ding{55}  & 25.3 / 7.2 & 91.1 / 97.3   \\
			& \ding{51} & \ding{51} & \textbf{21.9} / \textbf{3.6} & \textbf{92.0} / \textbf{98.7} \\
			\bottomrule
		\end{tabular}
        }
	\end{table}
}

\newcommand{\tableAblationNegLabelSelection}{
	\begin{table}[tb]
		\centering
		\caption{Ablation study on the type and the number of the negative labels. The ID dataset is VOC.}
		\label{tab:table_ablation_neglabel_selection}
		\scalebox{0.9}{
		\setlength{\tabcolsep}{0.5mm}
		\begin{tabular}{c|c|cccc}
			\toprule
			& & \multicolumn{4}{c}{OOD Dataset} \\
			\multirow{2}{*}{Negative Label} & \multirow{2}{*}{Number} & \multicolumn{2}{c}{MS-COCO} & \multicolumn{2}{c}{OpenImages} \\
			\cmidrule(lr){3-4}
			\cmidrule(lr){5-6}
			&   & FPR95$\downarrow$ & AUROC$\uparrow$  & FPR95$\downarrow$ & AUROC$\uparrow$\\
			\midrule
			``object'' &  1  & 13.9   & 96.0    &  45.3  & 82.9   \\
			\midrule
			\multirow{5}{*}{LLM\cite{hurst2024gpt4o}} &  1  & 12.9   & 96.2    &  42.3  & 85.1   \\
			&  5      & 12.4   & 95.5    &  38.7  & 86.0   \\
			&  10     & 13.6   & 84.9    &  40.6  & 85.2   \\
			&  20     &  \textbf{9.9}   & \textbf{95.9}    &  \textbf{23.3}  & \textbf{92.6}   \\
			&  30     & 15.5   & 93.8    &  38.2  & 87.4   \\
			\midrule
			\multirow{5}{*}{Corpus~\cite{princeton2010wordnet}} &  1  & 12.7   & 96.1    &  45.5  & 83.0   \\
			&  5      & 12.2   & 96.3    &  44.1  & 83.7   \\
			&  10     & 13.7   & 95.8    &  45.0  & 84.2   \\
			&  20     & \textbf{11.7}   & \textbf{96.6}    &  \textbf{42.4}  & \textbf{85.0}   \\
			&  30     & 12.0   & 95.8    &  40.8  & 84.6   \\
			\bottomrule
		\end{tabular}
	}
	\end{table}
}

\newcommand{\tableAblationLLMRobustness}{
\begin{table}[tb]
	\centering
	\caption{Robustness of NegAS to LLM choice, prompt design, and sampling randomness. Results on VOC (ID) with MS-COCO (OOD). Each experiment in (b)(c) is repeated three times. Qwen3 is an open-source LLM and reaches performance comparable to closed-source models.}
	\label{tab:table_ablation_llm_robustness}
	\scalebox{0.82}{
	\setlength{\tabcolsep}{1.2mm}
	\begin{tabular}{l|cc||l|cc||c|cc}
		\toprule
		\multicolumn{3}{c||}{(a) LLM Choice} & \multicolumn{3}{c||}{(b) Prompt Design} & \multicolumn{3}{c}{(c) Sampling (T / Top-p)} \\
		\midrule
		LLM & FPR95$\downarrow$ & AUROC$\uparrow$ & Prompt & FPR95$\downarrow$ & AUROC$\uparrow$ & T / Top-p & FPR95$\downarrow$ & AUROC$\uparrow$ \\
		\midrule
		GPT5     & 10.9 & 95.9 & Simple  & 10.4$\pm$0.5 & 95.8$\pm$0.6 & 0.2 / 0.8 & 11.0$\pm$0.4 & 95.0$\pm$0.4 \\
		GPT5.2   & 10.4 & 96.1 & Complex & 9.9$\pm$0.3  & 95.9$\pm$0.4 & 1.0 / 1.0 & 9.9$\pm$0.4  & 95.9$\pm$0.3 \\
		Opus4.5  & 11.0 & 95.2 &         &              &              & 1.2 / 1.0 & 10.7$\pm$0.5 & 96.0$\pm$0.2 \\
		Sonnet4  & 10.4 & 95.6 &         &              &              &           &              &  \\
		Qwen3 (open)  & 9.9 & 95.8 &         &              &              &           &              &  \\
		\bottomrule
	\end{tabular}
	}
\end{table}
}

\newcommand{\tableAblationStdAlphaMask}{
\begin{table}[htb]
    \centering
    \caption{Ablation on Gaussian noise standard deviation applied to negative text embeddings, OOD loss coefficient $\alpha$, and background mask. Results on VOC (ID) with MS-COCO / OpenImages (OOD).}
    \label{tab:table_ablation_abc}

    \hspace{-10mm}
    \begin{subtable}[t]{0.40\linewidth}
        \centering
        \scalebox{0.82}{
        \setlength{\tabcolsep}{1.2mm}
        \begin{tabular}{l|cc}
            \toprule
            Std & FPR95$\downarrow$ & AUROC$\uparrow$ \\
            \midrule
            0.1 & 15.7 / 41.9 & 94.4 / 87.0 \\
            0.01 & 11.9 / 36.0 & 95.3 / 86.9 \\
            0.05 & \textbf{9.9 / 23.3} & \textbf{95.9 / 92.6} \\
            0.0  & 9.9 / 33.4 & 96.7 / 89.5 \\
            \bottomrule
        \end{tabular}
        }
        \subcaption{Noise standard deviation.}
        \label{tab:ablation_std}
    \end{subtable}\hspace{-4mm}
    \begin{subtable}[t]{0.28\linewidth}
        \centering
        \scalebox{0.82}{
        \setlength{\tabcolsep}{1.2mm}
        \begin{tabular}{l|cc}
            \toprule
            $\alpha$ & FPR95$\downarrow$ & AUROC$\uparrow$ \\
            \midrule
            2 & 11.3 & 95.1 \\
            3 & \textbf{9.9} & \textbf{95.9} \\
            4 & 12.2 & 95.0 \\
              &      &      \\
            \bottomrule
        \end{tabular}
        }
        \subcaption{OOD loss coefficient $\alpha$.}
        \label{tab:ablation_alpha}
    \end{subtable}
    \begin{subtable}[t]{0.30\linewidth}
        \centering
        \scalebox{0.85}{
        \setlength{\tabcolsep}{1.2mm}
        \begin{tabular}{c|cc}
            \toprule
            Mask & FPR95$\downarrow$ & AUROC$\uparrow$   \\
            \midrule
            \ding{51} & \textbf{9.9} / \textbf{23.3} & \textbf{95.9} /  \textbf{92.6} \\
            \ding{55} & 10.9 / 32.2 & 95.4 / 88.8  \\
            &&\\
            &&\\
            \bottomrule
        \end{tabular}
        }
        \subcaption{Background mask.}
        \label{tab:ablation_mask}
    \end{subtable}

\end{table}
}

\newcommand{\tableHeadToHead}{
\begin{table}[tb]
    \centering
    \caption{Head-to-head comparison of OOD scoring functions ported onto YOLO-World. CLIP-classification OOD scores MCM~\cite{ming2022delving} and NegLabel~\cite{jiang2024negative} are computed on each proposal and contrasted with our NegS. ID dataset: VOC. ``YW'' is YOLO-World, ``$\text{YW}_{CoOp}$'' adds CoOp prompts. Best result per column is in \textbf{bold}.}
    \label{tab:table_head_to_head}
    \scalebox{0.82}{
    \setlength{\tabcolsep}{1.2mm}
    \begin{tabular}{l|cc|cc}
        \toprule
        \multirow{2}{*}{Method} & \multicolumn{2}{c|}{MS-COCO} & \multicolumn{2}{c}{OpenImages} \\
        \cmidrule(lr){2-3} \cmidrule(lr){4-5}
        & FPR95$\downarrow$ & AUROC$\uparrow$ & FPR95$\downarrow$ & AUROC$\uparrow$ \\
        \midrule
        YW + MCM      & 32.3 & 86.4 & 56.3 & 75.7 \\
        YW + NegLabel & 18.2 & 91.3 & 45.8 & 82.1 \\
        YW + NegS     & 13.5 & 94.7 & 38.8 & 83.6 \\
        \midrule
        $\text{YW}_{CoOp}$ + MCM      & 19.8 & 87.8 & 52.7 & 76.3 \\
        $\text{YW}_{CoOp}$ + NegLabel & 13.3 & 94.5 & 42.7 & 87.4 \\
        $\text{YW}_{CoOp}$ + NegS     & 11.4 & \textbf{96.2} & 33.3 & 88.8 \\
        \midrule
        YW + NegAS    & \textbf{9.9} & 95.9 & \textbf{23.3} & \textbf{92.6} \\
        \bottomrule
    \end{tabular}
    }
\end{table}
}

\newcommand{\tableIDmAPStd}{
\begin{table}[tb]
    \centering
    \caption{ID mAP with standard deviation across three random seeds. The gain over the YOLO-World baseline mainly stems from CoOp prompt tuning; on VOC the gap between $\text{YW}_{CoOp}$ and NegAS is within noise, so NegAS \emph{preserves} ID accuracy while substantially improving OOD detection.}
    \label{tab:table_id_map_std}
    \scalebox{0.9}{
    \setlength{\tabcolsep}{2mm}
    \begin{tabular}{l|cc}
        \toprule
        Method & VOC mAP (ID)$\uparrow$ & BDD mAP (ID)$\uparrow$ \\
        \midrule
        YW                  & 69.0 & 22.5 \\
        $\text{YW}_{CoOp}$  & 72.2$\pm$0.1 & 26.1$\pm$0.1 \\
        YW + NegAS          & 72.5$\pm$0.1 & 29.0$\pm$0.2 \\
        \bottomrule
    \end{tabular}
    }
\end{table}
}

\newcommand{\tableNegSim}{
\begin{table}[tb]
    \centering
    \caption{Mean cosine similarity between each OOD class embedding and its nearest neighbor in the selected negative-label set $\mathcal{Y}^{neg}$ (ID: VOC). The larger coverage gap on OpenImages explains why Gaussian noise on negative embeddings benefits OpenImages more than MS-COCO.}
    \label{tab:table_neg_sim}
    \scalebox{0.9}{
    \setlength{\tabcolsep}{2mm}
    \begin{tabular}{l|c}
        \toprule
        OOD set & Mean sim.\ to $\mathcal{Y}^{neg}$ \\
        \midrule
        MS-COCO    & 0.65 \\
        OpenImages & 0.40 \\
        \bottomrule
    \end{tabular}
    }
\end{table}
}

\usepackage{soul} 
\usepackage{booktabs} 
\usepackage{multirow} 
\usepackage[table]{xcolor} 
\usepackage{makecell} 
\usepackage{tikz} 
\usetikzlibrary{positioning, fit, shapes.misc, calc} 
\usepackage{pifont} 
\usepackage{booktabs}
\usepackage{multirow}
\usepackage{pifont}      
\usepackage{xcolor}      
\usepackage{subcaption}  
\usepackage{amsmath} 
\usepackage{tikz}

\usepackage[misc]{ifsym}

\definecolor{mycolor1}{HTML}{D5174E}
\definecolor{mycolor2}{HTML}{247AFD}
\definecolor{mycolor3}{HTML}{54b888}
\definecolor{mycolor4}{HTML}{c1c107}
\definecolor{mycolor5}{HTML}{44bb44}

\begin{document}

\title{NegAS: Negative Label Guided Attention and Scoring for Out-of-Distribution Object Detection with Vision Language Models} 

\titlerunning{NegAS: Negative Label Guided OOD Detection with VLMs}

\author{Yingjie Zhang\inst{1}\orcidlink{0009-0001-6952-2806} \and
Shuai Li\inst{1}\orcidlink{0000-0003-0760-5267}* \and
Peng Wang\inst{1}\orcidlink{0000-0001-7689-3405}
}

\authorrunning{Y. Zhang, S. Li and P. Wang.}


\institute{Northwestern Polytechnical University, Xi'An, China \\
\email{zyjzyj@mail.nwpu.edu.cn} \\
*: Corresponding Author
}

\maketitle

\begin{abstract}
Out-of-Distribution (OOD) detection is essential for ensuring the robustness and reliability of object detection systems deployed in safety-critical applications. While prior research has mainly focused on uni-modal detectors or vision-language model (VLM) based classifiers, the potential of VLM-based object detectors in OOD scenarios remains underexplored. In this work, we take the first step toward building OOD object detection methods upon VLMs. We identify two challenges specific to VLM detectors: (i) their text-guided attention enhances foreground with ID labels but treats background uniformly, leaving potential OOD regions unexploited for separating in-distribution (ID) from OOD instances; and (ii) their sigmoid-based multi-label outputs are incompatible with softmax-based OOD scores, calling for scoring functions consistent with VLM probabilistic outputs. Hence, we introduce Negative Label Guided Attention and Scoring (NegAS). To address (i), we propose a negative label guided attention module (NegA), where LLM-generated, visually-similar but semantically-different negative labels are used to guide attention toward potential OOD background regions. To address (ii), we introduce a novel sigmoid-based OOD scoring function (NegS) that leverages both ID and negative labels, producing strong responses for ID instances and suppressed responses for OOD ones. Extensive experiments demonstrate that our approach improves OOD detection performance by a large margin while maintaining ID accuracy, \eg, reducing the FPR95 by 11.4\% on the COCO dataset and 25.5\% on the OpenImages dataset compared to the baseline model. While initially designed for dense VLM detectors like YOLO-World, we successfully adapt NegAS to Grounding DINO, a query-based VLM transformer and achieve significant improvements, demonstrating the generalizability of our framework.

\end{abstract}

\section{Introduction}
\label{sec:intro}

Deep learning based object detection methods have achieved remarkable progresses in recent years. However, their accuracy and reliability often degrade in real-world scenarios. A critical issue arises when these models encounter objects of classes that were not present during training, known as Out-of-Distribution (OOD) data. In such cases, models tend to misclassify OOD objects as in-distribution (ID) ones, which can lead to severe consequences in safety-critical domains such as autonomous driving and medical image analysis. This makes OOD object detection (simply referred to as OOD detection) an essential research problem for enhancing the overall trustworthiness and reliability of deep learning detection methods.

OOD classification has been extensively studied in traditional uni-modal classification models, though such approaches generally do not incorporate textual information. Recently, vision-language models (VLMs), such as CLIP~\cite{radford2021learning}, which align textual and visual spaces, have demonstrated strong generalizability and are gradually becoming the mainstream for classification tasks. Consequently, OOD classification methods built upon CLIP have attracted significant attention. Existing research in this direction mainly explores how to exploit the alignment between text and vision to improve OOD classification performance, and can broadly be grouped into two categories. The first category, such as LoCoOp~\cite{miyai2023locoop}, NegPrompt~\cite{li2024learning}, LSN~\cite{nie2024out}, IDPrompt~\cite{bai2024id}, GalLoP ~\cite{lafon2024gallop}, focuses on designing negative prompts to enhance feature discriminability, to separate ID and OOD samples more effectively in the feature space. The second, such as MCM~\cite{ming2022delving}, GL-MCM~\cite{miyai2023zero}, NegLabel ~\cite{jiang2024negative}, EOE ~\cite{cao2024envisioning}, emphasizes developing improved OOD scoring functions that enable more accurate and reliable distinction between ID and OOD instances.

Motivated by the advances in OOD classification, the field of OOD detection has recently gained growing interest. Representative approaches—VOS~\cite{du2022vos}, STUD~\cite{du2022unknown}, SyncOOD~\cite{liu2024can}, WFS~\cite{wu2025percept}—largely follow the Faster R-CNN~\cite{ren2016faster} framework and utilize visual features only. More recently, vision language object detection models have demonstrated superior generalization compared to traditional uni-modal detectors. For example, YOLO-World~\cite{cheng2024yolo} aligns local image regions with text queries and achieves strong performance in downstream detection tasks. Nevertheless, its ability to address OOD detection remains largely unexplored. In this paper, we bridge this gap by investigating OOD object detection methods specially built upon VLMs. We emphasize that the two challenges below are not generic OOD issues but concrete obstacles that arise specifically when adapting a VLM object detector to OOD detection.

The first challenge is to better separate ID and OOD samples in the feature space.
YOLO-World introduces a text-guided attention layer that uses ID labels to emphasize foreground features and align vision–language representations. However, it treats all background regions uniformly. We argue that background regions vary in importance. We observe that certain background areas may contain potential OOD objects. Focusing more on these regions can yield a clearer decision boundary between ID and OOD instances. To this end, we propose a Negative label guided Attention (NegA) module that uses curated negative labels to reweight background features, guiding the network toward characteristics that contradict ID labels or resemble OOD labels. Besides, we design an LLM-based algorithm to generate high-quality negative labels that possess sufficient semantic differences from the ID labels, ensuring they provide meaningful and diverse guidance information.

The second challenge lies in designing an appropriate OOD scoring function. Most OOD methods are built on Faster R-CNN \cite{ren2016faster} trained with softmax-based losses, yielding softmax-based scores. In contrast, VLM detectors are trained with sigmoid-based losses, making a direct transfer suboptimal.
We address this with a Negative label guided Scoring (NegS) function tailored to VLM-based OOD object detection. Aligned with the probabilistic formulation of VLMs, it explicitly models the relative separation between ID and OOD samples, reducing overconfident predictions on unseen data.


We implement NegAS on two representative VLM detectors with different architectures—a dense detector YOLO-World and a query-based detector Grounding DINO—and conduct experiments on both. Results on these two baselines consistently validate the effectiveness of NegAS for OOD detection.



Our main contributions are as follows:
\begin{itemize}
	\item To the best of our knowledge, this is the first work to explore the potential of vision-language model (VLM) detectors for OOD object detection.
	\item We propose NegAS, which consists of a Negative label guided Attention (NegA) module and Scoring (NegS) function for VLM-based OOD object detection to enhance the separability between ID and OOD samples. Furthermore, we open-source our complete prompt templates and the comprehensive sets of curated negative labels to ensure full reproducibility.
    \item We implement NegAS on YOLO World, a dense VLM detector, and Grounding DINO, a query-based VLM detection transformer. Extensive experiments demonstrate the effectiveness of NegAS. This bridges the architectural gap between dense and query-based VLM detectors, demonstrating the generalizability of NegAS.
\end{itemize}

\section{Related Works}

\subsection{Pretrained vision-language models}
Thanks to the Transformer architecture ~\cite{vaswani2017attention}, pretrained vision-language models (VLMs) such as CLIP ~\cite{radford2021learning} have profoundly influenced vision tasks. Trained with contrastive learning, CLIP exhibits strong cross-modal alignment and generalization, becoming the de facto foundation for many vision-language models. Its robust representation ability has enabled a wide range of vision-language applications and motivated extensive research on adapting VLMs to downstream tasks.

\figureNegMining

\subsection{VLM object detectors}
Built upon pretrained VLMs, recent advances have enabled object detectors to move beyond closed-set taxonomies. ViLD ~\cite{gu2021openViLD} and GLIP ~\cite{li2022grounded} pioneered knowledge distillation and grounding-based approaches for open-vocabulary detection. Grounding DINO ~\cite{liu2024grounding} further integrates grounded pretraining into a transformer-based detector with language-guided query selection and cross-modality decoding, achieving state-of-the-art zero-shot performance. YOLO-World ~\cite{cheng2024yolo} extends the real-time YOLO framework with reparameterizable vision-language fusion and a prompt-then-detect paradigm, enabling practical open-vocabulary detection in real-world scenarios. These models highlight a trend toward tighter vision-language integration and stronger open-set generalization.

\subsection{Out-of-distribution object detection}

Early approaches attempt to synthesize OOD samples, either by perturbing in-distribution (ID) data or generating images with generative models. VOS ~\cite{du2022vos} perturbs ID distributions to create pseudo-OOD samples, training models with binary objectives to distinguish ID and OOD. SyncOOD ~\cite{liu2024can} feeds specially designed prompts into large language models (LLMs) to generate OOD labels, which are then used to edit images within bounding boxes via generative models. WFS ~\cite{wu2025percept} introduces the Open-Domain Unknown Object Detection (ODU-OD) task, targeting accurate detection of unknown objects across multiple unseen domains with diverse image styles. SIREN ~\cite{du2022siren} explicitly shapes feature spaces by enforcing compactness for ID categories and separability for OOD regions. These efforts highlight the challenge of extending detectors beyond the semantic space of their training data.

\subsection{Prompt learning}
Prompt learning, initially introduced in NLP, has recently been adapted to vision tasks as a flexible means of aligning VLMs with downstream objectives. CoOp ~\cite{zhou2022learning} adds learnable text prompts in CLIP, significantly improving its performance on image classification. CoCoOp ~\cite{zhou2022conditional} extends this idea with conditional prompts that generalize better to unseen classes. MaPLe ~\cite{khattak2023maple} introduces vision prompts to refine prompt tuning strategies to enable better vision-language alignment. In object detection, PromptDet ~\cite{feng2022promptdet} introduces task-specific prompts for open-vocabulary detection, showing the effectiveness of prompt learning in bridging the gap between vision and language. These advances make prompt learning a key technique for adapting large VLMs to diverse tasks.

\section{Methodology}

\figureMethodPipeline

We consider the closed-set fine-tuning setting for OOD detection: given a predefined set of ID categories, a dense VLM detector (\eg, YOLO-World) is fine-tuned on ID training data, and at test time must distinguish ID objects from OOD objects unseen during training.

To avoid confusion with related but distinct problems, we explicitly contrast three settings. \emph{(i) CLIP-based OOD classification} operates at the image level without localization, separating ID from OOD images on an ID--OOD mixed test set. \emph{(ii) Open-vocabulary object detection (OVOD)} detects objects from a predefined or open set of ID categories and is evaluated by detection accuracy on an ID-only test set; it has no notion of OOD, as every queried class is treated as ID. \emph{(iii) Our VLM-based OOD detection} fixes a finite ID label set (only ID categories are available at test time) and, on an ID--OOD mixed test set, assigns an OOD score to each detected box so as to separate ID from OOD objects. Evaluation thus focuses on ID/OOD separability rather than ID detection accuracy alone. In short, VLM-based OOD detection relates to VLM-based detection in the same way that CLIP-based OOD classification relates to CLIP-based classification.

As shown in \Cref{fig:oodattn}, our method is built upon YOLO-World ~\cite{cheng2024yolo}, a widely used VLM-based dense object detector. In this section, we first introduce how negative labels are generated in \cref{neg_label_selection}. Next, we explain how these negative labels guide the model to focus on important background regions in \cref{nega} and describe the training loss functions in \cref{training}. Finally, we present the NegS score computation based on negative labels in \cref{evaluation}.

\subsection{Negative label selection}
\label{neg_label_selection}

NegLabel~\cite{jiang2024negative} proposes to select negative labels from a large lexical database. However, we find that many words selected in this way are meaningless. To improve the quality of negative labels, we resort to large language models (LLMs) for negative label generation. Our mining pipeline differs from NegLabel's corpus mining in three concrete ways: (a) candidates are \emph{generated} by an LLM rather than retrieved from a fixed WordNet corpus; (b) we add a CLIP visual-similarity filter so that retained candidates are visually similar to ID objects; and (c) we rank candidates by semantic \emph{dissimilarity} to ID labels. A direct head-to-head against NegLabel-style corpus mining is reported in \Cref{tab:table_ablation_neglabel_selection} (the ``Corpus'' rows).
As illustrated in \Cref{fig:negmining}, given a collection of in-distribution (ID) category texts $\mathcal{Y}_{N}^{id}$ with $N$ entries, designed prompts are provided to the LLM (\eg, GPT-4o\cite{hurst2024gpt4o}) to produce a novel set of negative category texts, denoted as $\mathcal{C}^{neg}_{M}$ with $M$ entries. These negative texts describe objects that share certain visual characteristics with ID categories while differing semantically. Formally,
\begin{align}
	\mathcal{C}^{neg}_{M} &= \mathcal{LLM}(\mathcal{Y}^{id}_{N}; Prompts),
\end{align}

\noindent where $\mathcal{LLM}$ denotes the large language model. The used $Prompts$ and selection procedure are described below.

\textbf{Prompt design.} We provide the LLM with the following instructions.

System Prompts: \textit{``You are an AI visual assistant. You should not make up information or provide false details. The user will provide you with some In-Distribution (ID) class names in text, and you should respond with the corresponding OOD class names...''}. 

User Prompts: \textit{``The OOD class names should be of following features: 1. The OOD class names should be diverse and commonly seen in the real world... Now please list corresponding negative label names for ID names: [person, bird, ...]''}.

The complete prompt templates for each dataset and the full list of generated negative labels are provided in the supplementary materials.
We emphasize that negative-label generation is a \emph{one-time offline} step that is outside both training and inference: no LLM is queried during either phase. For VOC and BDD100K we release the full curated negative-label sets, so downstream users need zero LLM calls; for a new ID set, generation takes only seconds and a negligible API cost, and an open-source LLM (\eg, Qwen3) can be used as a drop-in alternative with comparable performance (\Cref{tab:table_ablation_llm_robustness}). Thus the use of a closed-source LLM does not limit the accessibility of NegAS.

\textbf{Negative label mining.}
Given the ID label set $\mathcal{Y}^{id}_N=\{y^{id}_j\}_{j=1}^{N}$, we prompt an LLM to generate a candidate pool
$\mathcal{C}=\{c_i\}_{i=1}^{|\mathcal{C}|}$.
Let $\mathcal{T}(\cdot)\in\mathbb{R}^{C}$ denote the text-encoder embedding and
$\mathrm{sim}(\mathbf{a},\mathbf{b})=\frac{\mathbf{a}^\top\mathbf{b}}{\|\mathbf{a}\|\|\mathbf{b}\|}$ be cosine similarity.

\emph{Visual similarity filtering.}
To ensure visual similarity to ID objects, we first build visual prototypes for each ID class from the ID dataset (\eg, by averaging CLIP~\cite{radford2021learning} image-encoder features of instances in that class).
Then, leveraging CLIP, we compute the image--text similarity between each candidate label $c_i$ and the set of ID prototypes.
We retain the $M$ candidates with a similarity over $\delta$ (default 0.5), forming a visually similar candidate set $\mathcal{C}^{neg}_M\subseteq\mathcal{C}$.

\emph{Semantic dissimilarity ranking.}
For each candidate $c_i\in\mathcal{C}^{neg}_M$, we measure its semantic closeness to the ID label set by the maximum text-text similarity:
\begin{equation}
s_i \;=\; \max_{1\le j\le N}\; \mathrm{sim}\!\big(\mathcal{T}(c_i),\, \mathcal{T}(y^{id}_j)\big).
\end{equation}
We define a dissimilarity score as $d_i \;=\; 1 - s_i, $
where dissimilarity score $d_i=0$ indicates that $c_i$ is identical (or extremely close) to some ID label in the text-embedding space.
Finally, inspired NegLabel~\cite{jiang2024negative}, we select the $K$ candidates with the largest dissimilarity score $d_i$ (equivalently, the smallest similarity $s_i$) to form the negative label set
$\mathcal{Y}^{neg}_K$, and discard candidates with large $s_i$ to reduce semantic overlap with ID labels.
This procedure yields negative labels that are visually similar while semantically distinct from the ID label set.

\subsection{Negative label guided attention}
\label{nega}

\textbf{Review of Text-Guided CSPLayer (YOLO-World).}
YOLO-World introduces Max-Sigmoid Attention (MSA) into CSPLayer to inject linguistic information into multi-scale visual features.
Let $T_N=\{t_j\}_{j=1}^{N}\in\mathbb{R}^{N\times C}=\mathcal{T}(\mathcal{Y}_N)$ be the text embeddings and
$I_l\in\mathbb{R}^{C\times H_l\times W_l}$ be the visual feature at FPN level $l\in\{3,4,5\}$.
For each spatial location $(m,n)$, denote $I_l^{(m,n)}\in\mathbb{R}^{C}$ as its feature vector. The Max-Sigmoid Attention $\hat{I}_l = MSA(I_l, T_N)$ is defined as:
\begin{align}
A_l^{(m,n)} &= \sigma\!\left(\max_{1\le j\le N}\; sim(t_j, I_l^{(m,n)})\right), \quad A_l\in\mathbb{R}^{H_l\times W_l}, \quad t_j \in \mathbb{R}^C, \\
\hat I_l &= I_l \odot A_l,
\end{align}
where $A_l^{(m,n)}$ is a scalar attention weight and $\odot$ denotes element-wise multiplication with broadcasting. Thus, regions closely associated with the texts are enhanced, while irrelevant regions are suppressed.

\textbf{YOLO-World with NegA (ours).}
Our goal is to leverage negative labels to \emph{amplify potential OOD evidence in background regions} while preserving the original ID detection capability of YOLO-World, thereby learning a more precise decision boundary between ID and OOD samples.
We impose two design principles:
(i) keep the YOLO-World architecture and parameters frozen to retain its generalization ability;
(ii) apply negative-label guidance \emph{only} on background regions to avoid conflicting with ID semantics.

\textbf{Dual-branch neck with shared parameters.}
As shown in \Cref{fig:oodattn}, we build two parallel neck branches with identical architecture and shared parameters.
The ID branch follows the original YOLO-World and produces ID-enhanced features, while the OOD branch is conditioned on negative text embeddings and produces OOD-enhanced features:
\begin{align}
\hat I_l^{id}  = MSA(I_l, T_N^{id}), \qquad
\hat I_l^{ood} = MSA(I_l, T_K^{neg}),
\end{align}
where $T_N^{id}=\mathcal{T}(\mathcal{Y}_N^{id})\in\mathbb{R}^{N\times C}$ and
$T_K^{neg}=\mathcal{T}(\mathcal{Y}_K^{neg})\in\mathbb{R}^{K\times C}$ are embeddings for ID and negative label sets, respectively.
Both $\hat I_l^{id}$ and $\hat I_l^{ood}$ are forwarded to the same detection head for classification, but the OOD supervision is restricted to background regions (details below).
We note that this design does not introduce gradient interference with the ID branch: the shared detector parameters are frozen and only the learnable prompts are updated, so $\mathcal{L}_{\mathit{ood}\_cls}$ does not directly distort the ID visual features. Moreover, $\mathcal{L}_{\mathit{ood}\_cls}$ acts only on background regions, where the ID target is also $\mathbf{0}$; hence the ID and OOD losses push the same logits in the same direction and are aligned rather than conflicting.
To broaden the semantic coverage of the LLM-generated negative labels, we perturb negative embeddings by Gaussian noise at each forward pass (default $\mu=0$, $\sigma=0.05$), which acts as a semantic-coverage regularizer rather than merely preventing overfitting (analyzed in \Cref{sec:ablation}).

\textbf{Background mask construction.}
\label{bg_mask_construction}
We construct a binary background mask $M_l\in\{0,1\}^{H_l\times W_l}$ from ground-truth (GT) bounding boxes during training.
For each FPN level $l\in\{3,4,5\}$ with stride $\{8,16,32\}$, we project GT boxes onto the feature-map grid by dividing box coordinates by the stride.
For each grid location $(m,n)$, we set
$M_l^{(m,n)}=0$ if $(m,n)$ falls \emph{inside} any projected GT box (foreground),
and $M_l^{(m,n)}=1$ otherwise (background).
This mask depends only on GT annotations.
The construction is intentionally conservative: any cell inside a projected box is treated as foreground, preventing negative-label guidance from penalizing ID regions near object boundaries.

\textbf{Training vs.\ inference.}
The OOD branch serves as \emph{training-time background augmentation}: it modulates attention using negative labels and provides additional supervision to shape a clearer ID--OOD boundary.
During training, $\hat I_l^{ood}$ is used to compute background-only OOD loss, while $\hat I_l^{id}$ is used for standard detection.
\textbf{At inference, we discard the NegA (OOD) branch for feature extraction} and keep only the ID branch to produce detections.
\textbf{Negative text embeddings are still retained} and are used only for computing the test-time OOD score (NegS in \Cref{evaluation}) together with ID-branch features, introducing no additional inference overhead in the neck.
During inference, we compute per-box classification scores $s^{id}\in [0,1]^{N}$ and $\tilde {s}^{neg}\in [0,1]^{K}$ using ID-branch features and $(T_N^{id},T_K^{neg})$, then obtain the OOD score by \Cref{equ:negs}. The inference pipeline is provided in supplementary materials.

\subsection{Training objectives}
\label{training}

Both $\hat I_l^{id}$ and $\hat I_l^{ood}$ are forwarded through the same detection head to predict scores over the $N$ ID classes $\mathcal{Y}_{N}^{id}$.
Let the set of predictions from the two branches be $\mathcal{P}^{id}=\{P_i^{id}\}_{i \in \Omega}$ and $\mathcal{P}^{ood}=\{P_i^{ood}\}_{i \in \Omega}$, where $\Omega$ indexes all predictions across all FPN levels $l\in\{3,4,5\}$.
For index $i\in\Omega$, we define $P_i^{id}=(\mathit{bbox}_i^{id}, s_i^{id})$ and $P_i^{ood}=(\mathit{bbox}_i^{ood}, s_i^{ood})$, where $\mathit{bbox}_i\in\mathbb{R}^{4}$ denotes the bounding box coordinates and $s_i\in[0,1]^N$ denotes sigmoid probabilities over the $N$ ID classes.

Both branches share the same classifier; the difference lies in how their features are produced and how their losses are applied.

\textbf{Losses.}
Let $\mathit{target}_i\in [0,1]^{N}$ be the soft target probability for the $i^{th}$ prediction, generated by the TaskAlignedAssigner adopted in YOLO-World. For the ID branch, we compute the standard Binary Cross Entropy (BCE) classification loss between the soft targets and the predicted sigmoid probabilities:
\begin{align}
\mathcal{L}_{\mathit{id}\_cls} = \sum_{i \in \Omega} \mathrm{BCE}(s_i^{id}, \mathit{target}_i),
\end{align}

For the OOD branch, we apply the classification loss \emph{only} on background regions using the background mask constructed in \Cref{bg_mask_construction}. Since each prediction $P_i^{ood} \in \mathcal{P}^{ood}$ is generated from a specific scale level $l(i)$ and spatial grid location $(m(i),n(i))$, its corresponding binary mask value $m_i = M_{l(i)}^{(m(i),n(i))} \in \{0,1\}$. The OOD classification loss is formulated as:
\begin{align}
\mathcal{L}_{\mathit{ood}\_cls} = \sum_{i \in \Omega} m_i \, \mathrm{BCE}(s_i^{ood}, \mathit{target}_i).
\end{align}
This formulation masks out foreground predictions ($m_i=0$), ensuring that only background predictions ($m_i=1$) contribute to the OOD loss.

\textbf{Total objective.}
The overall training objective is defined as:
\begin{align}
\mathcal{L} = \mathcal{L}_{\mathit{det}} + \alpha\,\mathcal{L}_{\mathit{ood}\_cls}, \qquad
\mathcal{L}_{\mathit{det}} = \mathcal{L}_{\mathit{id}\_cls} + \mathcal{L}_{\mathit{box}},
\end{align}
where $\alpha$ is a balancing coefficient (set to 3 by default), and $\mathcal{L}_{\mathit{box}}$ represents the standard bounding box regression loss computed exclusively on the ID branch.

\subsection{Test-time OOD detection}
\tableComparisonBaselines

\label{evaluation}
\textbf{Negative label guided OOD scoring (NegS).}
Existing OOD detectors often employ energy-based scores defined on softmax logits~\cite{du2022vos, liu2020energy}. For example, the energy used in VOS~\cite{du2022vos} is defined as:
\begin{align}
	E(P^{id}) = - \log \sum_{j=1}^{N} e^{z[j]}, \quad
	S(P^{id}) = \frac{e^{-E(P^{id})}}{1 + e^{-E(P^{id})}},
\end{align}
where $z[j]$ is the predicted logits for ID class $j$ prior to the softmax activation; $S(P^{id})$ is used as the final OOD score. 

However, traditional energy scores are theoretically derived from the softmax partition function (\ie, the denominator $\sum e^{z[j]}$), which assumes mutually exclusive classes. In contrast, open-vocabulary VLM detectors typically employ independent \emph{sigmoid} activations for multi-label classification. Consequently, applying softmax-based energy formulations to sigmoid logits is not directly compatible. More importantly, conventional energy scores do not exploit the negative-label semantics that we explicitly inject during training.

We therefore propose NegS, a sigmoid-based OOD scoring function assisted by negative labels.
Given ID and negative label sets $\mathcal{Y}_N^{id}$ and $\mathcal{Y}_K^{neg}$, we compute their respective text embeddings
$T_N^{id}=\mathcal{T}(\mathcal{Y}_N^{id})$ and $T_K^{neg}=\mathcal{T}(\mathcal{Y}_K^{neg})$.
For each predicted box $P^{id}=(\mathit{bbox}^{id}, s^{id}) \in \mathcal{P}^{id}$ produced by the \emph{ID branch},
we additionally compute negative-label sigmoid scores $\tilde{s}^{neg} \in [0,1]^{K}$ using the same ID-branch visual features and $T_K^{neg}$. NegS is defined as:
\begin{align}
\label{equ:negs}
NegS(P^{id}) = \max_{1 \leq j \leq N} s^{id}[j] \;-\; \max_{1 \leq k \leq K} \tilde{s}^{neg}[k].
\end{align}
A prediction strongly aligned with ID semantics yields a high first term, while a prediction aligned with negative semantics yields a high second term. Thus, NegS effectively separates ID objects from OOD or background regions by explicitly modeling their relative separation.

\section{Experiments}
\label{sec:experiments}

\subsection{Implementation details.}

\textbf{Datasets.}
\label{sec:datasets}
Following the experimental setting in~\cite{du2022vos}, we employ PASCAL VOC~\cite{everingham2010pascal} and BDD100K~\cite{yu2020bdd100k} as the In-Distribution (ID) datasets, containing 20 and 10 categories, respectively. For the Out-of-Distribution (OOD) evaluation, we adopt MS-COCO~\cite{lin2014microsoft} and OpenImages~\cite{kuznetsova2020open}, with overlapping categories removed to ensure no intersection with the ID datasets. 

\noindent
\textbf{Experimental settings.}
\label{sec:impl_details}
Our method is based on YOLO World~\cite{cheng2024yolo}, a recent open-vocabulary detection framework. For fair comparison, all models are trained under the same conditions unless otherwise specified.

We evaluate the following configurations:
\begin{itemize}
	\item $\text{YOLO World}$: The official YOLO World model without any additional training or modification.
	\item $\text{YOLO World}_{CoOp}$: YOLO World equipped with learnable text prompts following CoOp~\cite{zhou2022learning}, trained on two ID datasets using ID labels while keeping all other parameters frozen.
	\item NegAS (ours): The proposed Negative Label Guided Attention and Scoring method, trained on ID datasets using both ID and curated negative labels.
\end{itemize}

During training, with all other parameters except the prompts frozen, all models are trained using the AdamW optimizer with a learning rate of 1e-2 for 30 epochs with batch size 160 on 4 NVIDIA V100 GPUs. The image resolution is set to $640\times640$. For $\text{YOLO World}_{CoOp}$ and NegAS, we use prompt initialization ``\textit{a photo of}'', totally 3 learnable prompts, each with dimension 512. We select 20 negative labels from over 200 LLM-generated candidates for VOC, and 20 from over 300 candidates for BDD.

\noindent
\textbf{Evaluation metrics.}
\label{sec:eval_metrics}
Following standard OOD detection practice established by prior works such as VOS~\cite{du2022vos}, we report three metrics under the same evaluation pipeline:
FPR95 ($\downarrow$): False Positive Rate at 95\% True Positive Rate;
AUROC ($\uparrow$): Area Under the Receiver Operating Characteristic curve; and
mAP (ID) ($\uparrow$): Mean Average Precision on in-distribution data.

\subsection{Comparison with baselines and existing methods}
\label{sec:comp_baseline}
We first compare NegAS with two YOLO-World based baselines: the original model without fine-tuning and the CoOp-augmented variant trained with learnable prompts. As shown in \Cref{tab:table_comparison_baselines}, NegAS consistently surpasses both baselines across all evaluation metrics on VOC and BDD datasets, while maintaining competitive in-distribution accuracy.
On the PASCAL-VOC in-distribution dataset, NegAS achieves an FPR95 reduction from 48.8\% to 23.3\% when the OOD dataset is OpenImages, corresponding to a relative improvement of nearly 25\%, and raises the AUROC from 81.4\% to 92.6\%. On OOD dataset MS-COCO, NegAS yields the lowest FPR95 (9.9\%) and the highest AUROC (95.9\%), outperforming both YOLO-World and its CoOp-enhanced variant. Similarly, on the BDD dataset, NegAS also achieves the best results.

\tableIDmAPStd
We further clarify the source of the ID mAP gain. As reported in \Cref{tab:table_id_map_std} with standard deviations over three seeds, the improvement over the YOLO-World baseline (\eg, $69.0\!\rightarrow\!72.5$ on VOC) is largely attributable to CoOp prompt tuning; on VOC the gap between $\text{YW}_{CoOp}$ ($72.2\pm0.1$) and NegAS ($72.5\pm0.1$) is within noise. We therefore do \emph{not} claim that NegAS improves ID detection; rather, NegAS \emph{preserves} ID accuracy (with a stable gain on BDD) while delivering large OOD detection improvements, consistent with the OOD branch being discarded at inference.

\subsection{Generalization to other VLM detectors}
\label{sec:generalization}

To demonstrate the architectural generalizability of NegAS, we extend our framework to Grounding DINO, a query-based VLM detection transformer. Specifically, we introduce an auxiliary OOD branch parallel to the original architecture, as illustrated in supplementary materials.

During training, the model is optimized using AdamW with a learning rate of 1e-4 for 20 epochs. Consistent with our YOLO-World implementation, learnable prompts are initialized as ``a photo of'', and Gaussian noise ($\sigma = 0.05$) is injected into the text embeddings. During inference, the auxiliary OOD branch is discarded to maintain efficiency. The negative labels bypass feature fusion entirely and are used only with ID-label enhanced queries to compute NegS.

Unlike dense detectors, query-based models like Grounding DINO lack explicit absolute spatial information, making background mask construction complex. 
The detailed background mask construction is provided in the supplementary materials.
As shown in \Cref{tab:table_comparison_baselines}, NegAS significantly improves the OOD detection performance of Grounding DINO, proving that our negative-label-guided framework effectively generalizes well beyond dense VLM detectors.

\tableHeadToHead

\subsection{Comparison with CLIP-based OOD scoring functions}
\label{sec:comp_clip_scoring}
A natural question is whether the gains of NegAS simply come from porting existing CLIP-classification OOD techniques to a detector. To answer this, we directly transfer two representative CLIP-based OOD scoring functions, MCM~\cite{ming2022delving} and NegLabel~\cite{jiang2024negative}, onto each proposal of YOLO-World, and compare them with our NegS under the same backbone and prompts.
As shown in \Cref{tab:table_head_to_head}, NegS consistently outperforms both MCM- and NegLabel-style scoring, whether applied on the plain YOLO-World or on its CoOp variant. This indicates that the advantage of NegAS does \emph{not} stem from merely transferring VLM-classification OOD techniques: (i) NegA learns detector-specific OOD-aware features that scoring-only baselines cannot obtain, and (ii) NegS is explicitly tailored to the sigmoid-based outputs of VLM detectors. The full NegAS (NegA\,+\,NegS) further widens the gap, reaching the best FPR95 on both OOD sets.

\tableAblationComparison

\subsection{Ablation study}
\label{sec:ablation}
\textbf{Effectiveness of NegA and NegS.}
To better understand the contribution of each component, we conduct ablation experiments summarized in \Cref{tab:table_ablation_comparison}.
All variants are trained with learnable prompts for a consistent comparison.
Applying NegS consistently improves both AUROC and FPR95, confirming that incorporating negative semantics into the scoring function leads to more discriminative confidence estimation. 
The NegA module further enhances OOD separation, as it leverages negative label guidance to excavate latent OOD information. 
With both components combined, NegAS surpasses YOLO-World with learnable prompts by a large margin, validating the complementary effect of NegA and NegS in enhancing OOD detection robustness.

\tableAblationLLMRobustness  
\tableAblationStdAlphaMask
\textbf{Selection of negative labels.}
\tableAblationNegLabelSelection
In \Cref{tab:table_ablation_neglabel_selection}, we ablate the generation strategy and the number of negative labels. ``object'' means that the negative labels contain only a single ``object'' category.  ``LLM'' means that we prompt a large language model to generate potential negative labels given the ID labels. ``Corpus'' means that we select negative labels from a large-scale corpus database~\cite{princeton2010wordnet}, following the same procedure as in \cite{jiang2024negative}.
The results show that using a single ``object'' already outperforms the baseline. Labels selected from the corpus further improve the performance, but are still inferior to those chosen via the LLM. Notably, the ``Corpus'' rows reproduce NegLabel~\cite{jiang2024negative}'s lexical-database mining on YOLO-World and thus serve as a direct head-to-head comparison: at $K=20$, our LLM-mined labels achieve 9.9 vs.\ 11.7 FPR95 on VOC$\rightarrow$MS-COCO and 23.3 vs.\ 42.4 FPR95 on VOC$\rightarrow$OpenImages, confirming the benefit of our LLM-based mining over NegLabel-style corpus mining.

\textbf{Robustness to LLM choice and generation randomness.}
We evaluate the sensitivity of NegAS to LLM selection, prompt formulation, and sampling parameters on VOC (ID) with MS-COCO (OOD). As shown in \Cref{tab:table_ablation_llm_robustness}, NegAS achieves consistent performance across different LLMs, two prompt designs, and three temperature/top-p settings. Notably, the open-source Qwen3 reaches performance (FPR95 9.9, AUROC 95.8) on par with closed-source models, showing that NegAS does not depend on any particular proprietary LLM. The standard deviations across three repeated runs remain low, confirming that our method is robust to the choice of LLM and generation randomness.


\tableNegSim
\textbf{Influence of negative label embedding noise and OOD loss factor.}
We investigate the influence of presence and absence of noise on negative label embeddings and the OOD loss factor in \Cref{tab:ablation_std,tab:ablation_alpha}. Adding a slight noise improves OOD detection, and the effect is markedly stronger on VOC$\rightarrow$OpenImages than on VOC$\rightarrow$MS-COCO. We attribute this asymmetry not to overfitting but to the \emph{semantic coverage} of the selected negative labels. As shown in \Cref{tab:table_neg_sim}, the negative labels are farther from the OpenImages OOD classes (mean cosine similarity 0.40) than from the MS-COCO OOD classes (0.65), \ie, a larger coverage gap on OpenImages. The Gaussian noise locally expands the negative-label embedding neighborhood, which helps close this gap on OpenImages but brings little benefit on MS-COCO where the labels already provide sufficient coverage. We observe that the OOD loss is very small. Considering that an appropriate amount of OOD correction is beneficial—insufficient correction fails to take effect, while excessive correction may disturb the original feature distribution—we investigate the influence of the OOD loss coefficient $\alpha$. Empirically, when $\alpha = 3$, the method achieves the best performance.

\figureHeatmap

\textbf{Background mask.}
As shown in \Cref{tab:ablation_mask}, calculating OOD loss only on background regions outperforms calculating the overall image, confirming that foreground features modulated by ID and negative labels can conflict.

\subsection{Qualitative results}
\label{sec:qualitative_results}
We visualized feature heatmaps in \Cref{fig:heatmap_visualization} to illustrate the attention shift introduced by NegA.
The ID label enhanced feature maps show high responses on ID classes (chair, TV, bird, dining table), while negative label guided attentions focus on potential OOD regions (doll, desk, branch, bread). This demonstrates that NegAS guides the model toward latent OOD information, helping to better distinguish ID and OOD samples.
Red regions indicate higher attention weights.
The ID label enhanced features (middle row) highlight ID object regions (\eg, chair, TV, bird, dining table), while the negative label guided features (bottom row) focus on potential OOD regions (\eg, doll, desk, branch, bread), demonstrating NegA's ability to attend to OOD-relevant background areas.

\section{Conclusion}

In this paper, we explored the OOD problem for dense vision-language detection models. First, we proposed Negative Label Guided Attention (NegA), which uses negative labels generated by an LLM to guide the model to focus more on important OOD background regions. This enhances feature diversity and learns a better decision boundary through prompt learning. Second, we introduced Negative Label Assisted OOD Scoring (NegS) function during the testing phase. This function is specifically designed for sigmoid-based VLMs and enables better differentiation between ID and OOD samples. Finally, we generalized the proposed NegAS to a query-based VLM detector to investigate the generalizalibity.
Extensive experiments on multiple benchmarks demonstrated the effectiveness and robustness of NegAS and its generalizability to query-based VLM detectors. 


\par\vfill\par

\bibliographystyle{splncs04}
\bibliography{main}

@inproceedings{radford2021learning,
	title={Learning transferable visual models from natural language supervision},
	author={Radford, Alec and Kim, Jong Wook and Hallacy, Chris and Ramesh, Aditya and Goh, Gabriel and Agarwal, Sandhini and Sastry, Girish and Askell, Amanda and Mishkin, Pamela and Clark, Jack and others},
	booktitle={International conference on machine learning},
	pages={8748--8763},
	year={2021},
	organization={PmLR}
}

@inproceedings{li2022grounded,
	title={Grounded language-image pre-training},
	author={Li, Liunian Harold and Zhang, Pengchuan and Zhang, Haotian and Yang, Jianwei and Li, Chunyuan and Zhong, Yiwu and Wang, Lijuan and Yuan, Lu and Zhang, Lei and Hwang, Jenq-Neng and others},
	booktitle={Proceedings of the IEEE/CVF conference on computer vision and pattern recognition},
	pages={10965--10975},
	year={2022}
}

@article{ming2022delving,
	title={Delving into out-of-distribution detection with vision-language representations},
	author={Ming, Yifei and Cai, Ziyang and Gu, Jiuxiang and Sun, Yiyou and Li, Wei and Li, Yixuan},
	journal={Advances in neural information processing systems},
	volume={35},
	pages={35087--35102},
	year={2022}
}

@article{miyai2023zero,
	title={Zero-shot in-distribution detection in multi-object settings using vision-language foundation models},
	author={Miyai, Atsuyuki and Yu, Qing and Irie, Go and Aizawa, Kiyoharu},
	journal={arXiv preprint arXiv:2304.04521},
	year={2023}
}

@article{jiang2024negative,
	title={Negative label guided ood detection with pretrained vision-language models},
	author={Jiang, Xue and Liu, Feng and Fang, Zhen and Chen, Hong and Liu, Tongliang and Zheng, Feng and Han, Bo},
	journal={arXiv preprint arXiv:2403.20078},
	year={2024}
}

@article{cao2024envisioning,
	title={Envisioning outlier exposure by large language models for out-of-distribution detection},
	author={Cao, Chentao and Zhong, Zhun and Zhou, Zhanke and Liu, Yang and Liu, Tongliang and Han, Bo},
	journal={arXiv preprint arXiv:2406.00806},
	year={2024}
}

@article{miyai2023locoop,
	title={Locoop: Few-shot out-of-distribution detection via prompt learning},
	author={Miyai, Atsuyuki and Yu, Qing and Irie, Go and Aizawa, Kiyoharu},
	journal={Advances in Neural Information Processing Systems},
	volume={36},
	pages={76298--76310},
	year={2023}
}

@inproceedings{li2024learning,
	title={Learning transferable negative prompts for out-of-distribution detection},
	author={Li, Tianqi and Pang, Guansong and Bai, Xiao and Miao, Wenjun and Zheng, Jin},
	booktitle={Proceedings of the IEEE/CVF Conference on Computer Vision and Pattern Recognition},
	pages={17584--17594},
	year={2024}
}

@inproceedings{lafon2024gallop,
	title={Gallop: Learning global and local prompts for vision-language models},
	author={Lafon, Marc and Ramzi, Elias and Rambour, Cl{\'e}ment and Audebert, Nicolas and Thome, Nicolas},
	booktitle={European Conference on Computer Vision},
	pages={264--282},
	year={2024},
	organization={Springer}
}

@inproceedings{nie2024out,
	title={Out-of-distribution detection with negative prompts},
	author={Nie, Jun and Zhang, Yonggang and Fang, Zhen and Liu, Tongliang and Han, Bo and Tian, Xinmei},
	booktitle={The twelfth international conference on learning representations},
	year={2024}
}

@inproceedings{bai2024id,
	title={Id-like prompt learning for few-shot out-of-distribution detection},
	author={Bai, Yichen and Han, Zongbo and Cao, Bing and Jiang, Xiaoheng and Hu, Qinghua and Zhang, Changqing},
	booktitle={Proceedings of the IEEE/CVF Conference on Computer Vision and Pattern Recognition},
	pages={17480--17489},
	year={2024}
}

@inproceedings{liu2024grounding,
	title={Grounding dino: Marrying dino with grounded pre-training for open-set object detection},
	author={Liu, Shilong and Zeng, Zhaoyang and Ren, Tianhe and Li, Feng and Zhang, Hao and Yang, Jie and Jiang, Qing and Li, Chunyuan and Yang, Jianwei and Su, Hang and others},
	booktitle={European conference on computer vision},
	pages={38--55},
	year={2024},
	organization={Springer}
}

@inproceedings{cheng2024yolo,
	title={Yolo-world: Real-time open-vocabulary object detection},
	author={Cheng, Tianheng and Song, Lin and Ge, Yixiao and Liu, Wenyu and Wang, Xinggang and Shan, Ying},
	booktitle={Proceedings of the IEEE/CVF conference on computer vision and pattern recognition},
	pages={16901--16911},
	year={2024}
}

@article{du2022vos,
	title={Vos: Learning what you don't know by virtual outlier synthesis},
	author={Du, Xuefeng and Wang, Zhaoning and Cai, Mu and Li, Yixuan},
	journal={arXiv preprint arXiv:2202.01197},
	year={2022}
}

@inproceedings{liu2024can,
	title={Can OOD Object Detectors Learn from Foundation Models?},
	author={Liu, Jiahui and Wen, Xin and Zhao, Shizhen and Chen, Yingxian and Qi, Xiaojuan},
	booktitle={European Conference on Computer Vision},
	pages={213--231},
	year={2024},
	organization={Springer}
}

@inproceedings{du2022unknown,
	title={Unknown-aware object detection: Learning what you don't know from videos in the wild},
	author={Du, Xuefeng and Wang, Xin and Gozum, Gabriel and Li, Yixuan},
	booktitle={Proceedings of the IEEE/CVF conference on computer vision and pattern recognition},
	pages={13678--13688},
	year={2022}
}

@inproceedings{wu2025percept,
	title={Percept, Memory, and Imagine: World Feature Simulating for Open-Domain Unknown Object Detection},
	author={Wu, Aming and Deng, Cheng},
	booktitle={Proceedings of the Computer Vision and Pattern Recognition Conference},
	pages={4682--4691},
	year={2025}
}

@article{vaswani2017attention,
	title={Attention is all you need},
	author={Vaswani, Ashish and Shazeer, Noam and Parmar, Niki and Uszkoreit, Jakob and Jones, Llion and Gomez, Aidan N and Kaiser, {\L}ukasz and Polosukhin, Illia},
	journal={Advances in neural information processing systems},
	volume={30},
	year={2017}
}

@article{gu2021openViLD,
	title={Open-vocabulary object detection via vision and language knowledge distillation},
	author={Gu, Xiuye and Lin, Tsung-Yi and Kuo, Weicheng and Cui, Yin},
	journal={arXiv preprint arXiv:2104.13921},
	year={2021}
}

@inproceedings{lin2014microsoft,
	title={Microsoft coco: Common objects in context},
	author={Lin, Tsung-Yi and Maire, Michael and Belongie, Serge and Hays, James and Perona, Pietro and Ramanan, Deva and Doll{\'a}r, Piotr and Zitnick, C Lawrence},
	booktitle={European conference on computer vision},
	pages={740--755},
	year={2014},
	organization={Springer}
}

@article{du2022siren,
	title={Siren: Shaping representations for detecting out-of-distribution objects},
	author={Du, Xuefeng and Gozum, Gabriel and Ming, Yifei and Li, Yixuan},
	journal={Advances in neural information processing systems},
	volume={35},
	pages={20434--20449},
	year={2022}
}

@article{zhou2022learning,
	title={Learning to prompt for vision-language models},
	author={Zhou, Kaiyang and Yang, Jingkang and Loy, Chen Change and Liu, Ziwei},
	journal={International Journal of Computer Vision},
	volume={130},
	number={9},
	pages={2337--2348},
	year={2022},
	publisher={Springer}
}

@inproceedings{yu2020bdd100k,
	title={Bdd100k: A diverse driving dataset for heterogeneous multitask learning},
	author={Yu, Fisher and Chen, Haofeng and Wang, Xin and Xian, Wenqi and Chen, Yingying and Liu, Fangchen and Madhavan, Vashisht and Darrell, Trevor},
	booktitle={Proceedings of the IEEE/CVF conference on computer vision and pattern recognition},
	pages={2636--2645},
	year={2020}
}

@article{everingham2010pascal,
	title={The pascal visual object classes (voc) challenge},
	author={Everingham, Mark and Van Gool, Luc and Williams, Christopher KI and Winn, John and Zisserman, Andrew},
	journal={International journal of computer vision},
	volume={88},
	number={2},
	pages={303--338},
	year={2010},
	publisher={Springer}
}

@article{kuznetsova2020open,
	title={The open images dataset v4: Unified image classification, object detection, and visual relationship detection at scale},
	author={Kuznetsova, Alina and Rom, Hassan and Alldrin, Neil and Uijlings, Jasper and Krasin, Ivan and Pont-Tuset, Jordi and Kamali, Shahab and Popov, Stefan and Malloci, Matteo and Kolesnikov, Alexander and others},
	journal={International journal of computer vision},
	volume={128},
	number={7},
	pages={1956--1981},
	year={2020},
	publisher={Springer}
}

@inproceedings{zhou2022conditional,
	title={Conditional prompt learning for vision-language models},
	author={Zhou, Kaiyang and Yang, Jingkang and Loy, Chen Change and Liu, Ziwei},
	booktitle={Proceedings of the IEEE/CVF conference on computer vision and pattern recognition},
	pages={16816--16825},
	year={2022}
}

@inproceedings{khattak2023maple,
	title={Maple: Multi-modal prompt learning},
	author={Khattak, Muhammad Uzair and Rasheed, Hanoona and Maaz, Muhammad and Khan, Salman and Khan, Fahad Shahbaz},
	booktitle={Proceedings of the IEEE/CVF conference on computer vision and pattern recognition},
	pages={19113--19122},
	year={2023}
}

@inproceedings{feng2022promptdet,
	title={Promptdet: Towards open-vocabulary detection using uncurated images},
	author={Feng, Chengjian and Zhong, Yujie and Jie, Zequn and Chu, Xiangxiang and Ren, Haibing and Wei, Xiaolin and Xie, Weidi and Ma, Lin},
	booktitle={European conference on computer vision},
	pages={701--717},
	year={2022},
	organization={Springer}
}

@article{liu2020energy,
	title={Energy-based out-of-distribution detection},
	author={Liu, Weitang and Wang, Xiaoyun and Owens, John and Li, Yixuan},
	journal={Advances in neural information processing systems},
	volume={33},
	pages={21464--21475},
	year={2020}
}

@inproceedings{sastry2020detecting,
	title={Detecting out-of-distribution examples with gram matrices},
	author={Sastry, Chandramouli Shama and Oore, Sageev},
	booktitle={International conference on machine learning},
	pages={8491--8501},
	year={2020},
	organization={PMLR}
}

@inproceedings{hsu2020generalized,
	title={Generalized odin: Detecting out-of-distribution image without learning from out-of-distribution data},
	author={Hsu, Yen-Chang and Shen, Yilin and Jin, Hongxia and Kira, Zsolt},
	booktitle={Proceedings of the IEEE/CVF conference on computer vision and pattern recognition},
	pages={10951--10960},
	year={2020}
}

@article{tack2020csi,
	title={Csi: Novelty detection via contrastive learning on distributionally shifted instances},
	author={Tack, Jihoon and Mo, Sangwoo and Jeong, Jongheon and Shin, Jinwoo},
	journal={Advances in neural information processing systems},
	volume={33},
	pages={11839--11852},
	year={2020}
}

@inproceedings{wilson2023safe,
	title={Safe: Sensitivity-aware features for out-of-distribution object detection},
	author={Wilson, Samuel and Fischer, Tobias and Dayoub, Feras and Miller, Dimity and S{\"u}nderhauf, Niko},
	booktitle={Proceedings of the ieee/cvf international conference on computer vision},
	pages={23565--23576},
	year={2023}
}

@inproceedings{wu2023deep,
	title={Deep feature deblurring diffusion for detecting out-of-distribution objects},
	author={Wu, Aming and Chen, Da and Deng, Cheng},
	booktitle={Proceedings of the IEEE/CVF international conference on computer vision},
	pages={13381--13391},
	year={2023}
}

@inproceedings{wu2023discriminating,
	title={Discriminating known from unknown objects via structure-enhanced recurrent variational autoencoder},
	author={Wu, Aming and Deng, Cheng},
	booktitle={Proceedings of the IEEE/CVF conference on computer vision and pattern recognition},
	pages={23956--23965},
	year={2023}
}

@misc{princeton2010wordnet,
	author       = {Princeton University},
	title        = {About WordNet},
	year         = {2010},
	howpublished = {\url{https://wordnet.princeton.edu/}},
	note         = {Accessed: 2025-11-09}
}

@article{hurst2024gpt4o,
  title={Gpt-4o system card},
  author={Hurst, Aaron and Lerer, Adam and Goucher, Adam P and Perelman, Adam and Ramesh, Aditya and Clark, Aidan and Ostrow, AJ and Welihinda, Akila and Hayes, Alan and Radford, Alec and others},
  journal={arXiv preprint arXiv:2410.21276},
  year={2024}
}

@article{ren2016faster,
  title={Faster R-CNN: Towards real-time object detection with region proposal networks},
  author={Ren, Shaoqing and He, Kaiming and Girshick, Ross and Sun, Jian},
  journal={IEEE transactions on pattern analysis and machine intelligence},
  volume={39},
  number={6},
  pages={1137--1149},
  year={2016},
  publisher={IEEE}
}


\end{document}